%% file: paper.tex
\documentclass{scrartcl}
\pdfoutput=1

\usepackage{mathptmx}       
\usepackage{helvet}         
\usepackage{courier}        
\usepackage{type1cm}        
\usepackage{makeidx}         
\usepackage{graphicx}        
\usepackage{multicol}        
\usepackage[bottom]{footmisc}

\usepackage{amsmath}
\usepackage{amsfonts}
\usepackage{bm}

\begin{document}
\title{Fuzzy Fibers:\\ Uncertainty in dMRI Tractography}
\author{Thomas Schultz\thanks{\quad University of Bonn, Germany. E-Mail:
    schultz@cs.uni-bonn.de}, Anna Vilanova\thanks{\quad Technical University
    Eindhoven, Netherlands. E-Mail:
    \{a.vilanova,r.brecheisen\}@tue.nl}, Ralph
  Brecheisen,\hspace{-1.5mm}\footnotemark[2]\; and Gordon
  Kindlmann\thanks{\quad University of Chicago, USA. E-Mail: glk@uchicago.edu}}

\maketitle

\begin{abstract}
  Fiber tracking based on diffusion weighted Magnetic Resonance
  Imaging (dMRI) allows for noninvasive reconstruction of fiber
  bundles in the human brain. In this chapter, we discuss sources of
  error and uncertainty in this technique, and review strategies that
  afford a more reliable interpretation of the results. This includes
  methods for computing and rendering probabilistic tractograms, which
  estimate precision in the face of measurement noise and
  artifacts. However, we also address aspects that have received less
  attention so far, such as model selection, partial voluming, and the
  impact of parameters, both in preprocessing and in fiber tracking
  itself. We conclude by giving impulses for future research.
\end{abstract}
\section{Introduction}
\label{sec:introduction}

\input{01-intro}

\section{Noise and Artifacts}
\label{sec:noise-artifacts}

\input{02-1-strategies}

\input{02-2-rendering}

\section{Other Factors}
\label{sec:other-factors}

\input{03-1-parameters}

\input{03-2-uncertainty}

\input{03-3-part-vol}

\section{Perspectives}
\label{sec:perspectives}

\input{04-1-evidence}

\input{04-2-reproducibility}

\section{Conclusion}
\label{sec:conclusion}

\input{05-conclusion}

\footnotesize
\bibliographystyle{plain}
\bibliography{schultz,kindlmann,vilanova}

\end{document}

%% file: 01-intro.tex
Diffusion weighted MRI (dMRI) is a modern variant of Magnetic
Resonance Imaging that allows for noninvasive, spatially
resolved measurement of apparent self-diffusion coefficients. Since
fibrous tissues such as nerve fiber bundles in the human brain
constrain water molecules such that they diffuse more freely
along fibers than orthogonal to them, the apparent diffusivity
depends on the direction of measurement, and allows us to infer the
main fiber direction.

Based on such data, tractography algorithms reconstruct the
trajectories of major nerve fiber bundles. The most classic variant is
streamline tractography, in which tracking starts at some seed point
and proceeds in small steps along the inferred direction. In its
simplest form, this results in one space curve per seed point. It has
been observed that many of the resulting streamlines agree with known
anatomy \cite{Catani:2002}. Tractography is also supported by
validation studies that have used software simulations, physical and
biological phantoms \cite{Hubbard:2009}.


Tractography is currently the only technique for noninvasive
reconstruction of fiber bundles in the human brain. This has created
much interest among neuroscientists, who are looking for evidence of
how connectivity between brain regions varies between different groups
of subjects \cite{Sotiropoulos:2013}, as well as neurosurgeons, who
would like to know the exact spatial extent of specific fiber bundles
in individual patients.

However, drawing reliable inference from dMRI is challenging. Even
though a randomized controlled study has shown that using dMRI in
cerebral glioma surgery reduces postoperative motor deficits and
increases survival times \cite{Wu:2007}, neurosurgeons have observed
that some methods for tractography underestimate the true size of
bundles \cite{Kinoshita:2005} and they are still unsatisfied with the
degree of reproducibility that is achieved with current software
packages \cite{Burgel:2009}.

In order to establish tractography as a reliable and widely accepted
technique, it is essential to gain a full understanding of its
inherent sources of error and uncertainty. It is the goal of this
chapter to give an introduction to these problems, to present existing
approaches that have tried to mitigate or model them, and to outline
some areas where more work is still needed.


%% file: 02-1-strategies.tex
\subsection{Strategies for Probabilistic Tractography}
\label{sec:prob-strategies}

It is the goal of probabilistic tractography to estimate the
variability in fiber bundle reconstructions that is due to measurement
noise. This is often referred to as precision of the reconstructed
bundle trajectory \cite{Jones:2010}. Due to additional types of error
in data acquisition and modeling, which will be covered later in this
chapter, it is not the same as accuracy (i.e., likelihood of a true
anatomical connection) \cite{Jones:2005}. Current approaches
do not account for factors such as repositioning of the head or
variations in scanner hardware over time, which further affect
repeatability in practice.

Rather than only inferring the most likely fiber direction,
probabilistic approaches derive a probability distribution of fiber
directions from the data. The first generation of probabilistic
tractography methods has done so by fitting the diffusion tensor model
to the data, and using the result to parameterize a probability
distribution in a heuristic manner. This often assumes that
the fiber distribution is related to a sharpened version of the
diffusivity profile \cite{Koch:2002}, sometimes regularized by a
deliberate bias towards the direction of the previous tracking step
\cite{Bjornemo:2002,Hagmann:2003}. Programmable graphics hardware
accelerates the sampling of such models, and enables immediate
visualization of the results \cite{McGraw:2007}. Parker et al.\
\cite{Parker:2003JMRI} present two different fiber distribution models
that are parameterized by measures of diffusion anisotropy. Subsequent
work allows for multimodal distributions that capture fiber crossings,
and uses the observed variation of principal eigenvectors in synthetic
data to calibrate model parameters \cite{Parker:2003}.

In contrast to these techniques, which use the model parameters from a
single fit, a second generation of probabilistic tractography methods
estimates the posterior distribution of fiber model parameters, based
on the full information from the measurements, which includes fitting
residuals.  Behrens et al.\ \cite{Behrens:2003} do so in an objective
Bayesian framework, which aims at making as few assumptions as possible,
by choosing noninformative priors. They have later extended the
``ball-and-stick'' model that underlies their framework to allow for
multiple fiber compartments \cite{Behrens:2007}.

Bootstrapping estimates the distribution of anisotropy measures
\cite{Pajevic:2003} or fiber directions \cite{Jones:2003,Schultz:2013}
by repeated model fitting, after resampling data from a limited number
of repeated scans. This has been used as the foundation of another
line of probabilistic tractography approaches
\cite{Jones:2005,Lazar:2005}. As an alternative to estimating the
amount of noise from repeated measurements, wild bootstrapping takes
its noise estimates from the residuals that remain when fitting a
model to a single set of measurements \cite{Whitcher:2008}. This has
been proposed as an alternative to repetition-based bootstrapping for
cases where only a single acquisition is available
\cite{Jones:2008}. Residual bootstrapping \cite{Chung:2006} builds on
the same basic idea, but allows for resampling residuals between
gradient directions, by modeling the heteroscedasticity in them. It
has not only been combined with the diffusion tensor model, but also
with constrained deconvolution, which allows for multiple fiber
tractography \cite{Jeurissen:2011}.


%% file: 02-2-rendering.tex
\subsection{Rendering Probabilistic Tractograms}
\label{sec:prob-rendering}

After estimating the reproducibility of white matter fiber tracts by one of the above-described methods, we can represent the results in one of two ways: voxel-centric or tract-centric.  The voxel-centric representation assigns scores to individual voxels, where each voxel stores the percentage of tracts passing through it. They represent the reproducibility with which a connection from one voxel position to the seeding region is inferred from the data. The resulting 3D volume data sets are sometimes called probability or confidence maps, and are often visualized by volume rendering techniques and 2D color maps \cite{McGraw:2007}.

Tract-centric techniques include the ConTrack algorithm~\cite{Sherbondy:2008}, which assigns a score to each generated tract. It reflects confidence in the pathway as a whole, based on its agreement with the data and assumptions on fiber length and smoothness. Ehricke et al.~\cite{Ehricke:2006} define a confidence score that varies along the fiber, and color code it on the streamline. Jones et al.~\cite{Jones:2005b,Jones:2008} use hyperstreamlines to visualize the variability of fiber tracts obtained using bootstrap or wild-bootstrap methods. They also demonstrate that using standard streamlines to render all fiber variations equally fails to give an impression of which fibers are stable and which are outliers.

Brecheisen et al.~\cite{Brecheisen:2011} propose illustrative confidence intervals where intervals are based on distances or pathway scores. Illustrative techniques, i.e., silhouette and outlines, are used to visualize these intervals. Interaction and Focus+Context widgets are used to extend the simplified illustrative renderings with more detailed information.

Schultz et al.~\cite{Schultz:Vis2007} cluster the voxels in which probabilistic tractography terminates, based on the seed points from which they are reached. They then derive a per-voxel score that indicates how frequently the voxel was involved in a connection between two given clusters. Fuzzy fiber bundle geometry is defined by isosurfaces of this score, with different isovalues representing different levels of precision.


%% file: 03-1-parameters.tex
\subsection{Impact of Parameters}
\label{sec:parameters}

One source of uncertainty in dMRI tractography that has not received much attention is parameter sensitivity. Most tractography algorithms depend on user-defined parameters, which results in a poor reproducibility of the output results. Some reproducibility studies for concrete applications have been reported~\cite{Ciccarelli:2003,Wakana:2007}. However, there does not exist an automatic solution that resolves the problem in a general manner. The stability of the parameter setting is relevant information for both neuroscientists and neurosurgeons who are trying to assess whether their fiber tracking results are stable. Visualization can play an important role to help this assessment.

\begin{figure}[h]
\centering
\includegraphics[width = \textwidth]{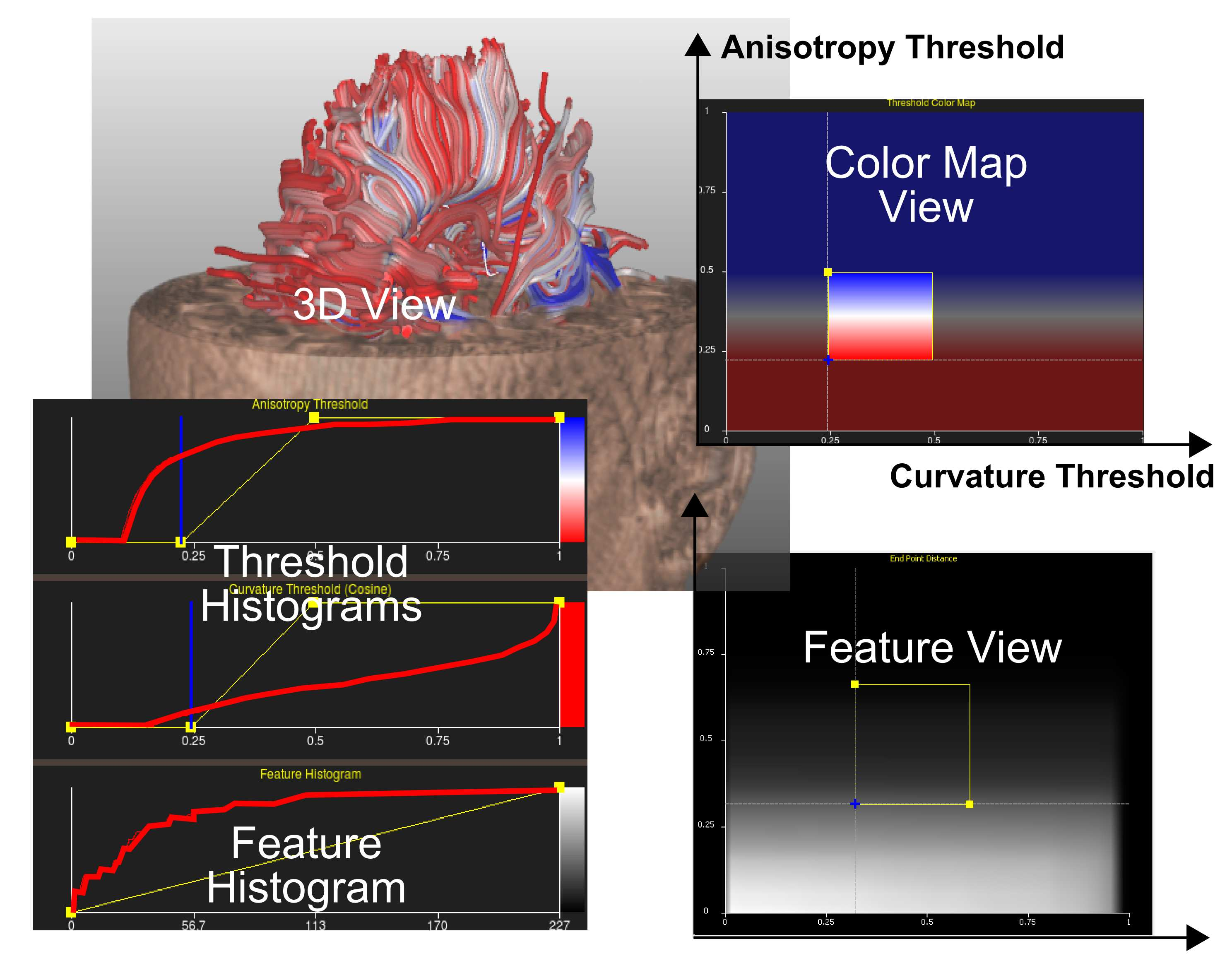}
\caption{Main viewports of Brecheisen et al.~\cite{Brecheisen:2009} exploration tool. Top-left: 3D visualization
of fiber tract together with anatomical context and axial fractional
anisotropy slice. Top-right: color map view used for selecting individual
threshold combinations and definition of color detail regions. Bottom-right:
feature map view showing changes in quantitative tract features
as a function of threshold combination at discrete sample points of the
parameter space. Bottom-left: cumulative histograms of both threshold
and feature values.}
\label{fig:parameterEstimation}
\end{figure}

Brecheisen et al.~\cite{Brecheisen:2009} build a parameter space by sampling combinations of stopping criteria for DTI streamline tractography. Stopping criteria primarily affect fiber length. The investigation of parameter sensitivity is based on generating a streamline set that covers the whole parameter space of stopping criteria. Afterwards, selective culling is performed to display specific streamline collections from the parameter space. This is done by selecting parameter combinations using 2D widgets such as the feature histogram displayed in Figure~\ref{fig:parameterEstimation}. An example feature is average fiber density per voxel. These views help the user to identify stable parameter settings, thereby improving the ability to compare groups of subjects based on quantitative tract features.

Jiao et al.~\cite{Jiao:2010} introduce a toolkit based on three streamline distances that are used to measure  differences between fiber bundles. The user can vary parameters that affect the results of the fiber tractography and measure the resulting differences based on these distances. This allows them to quantify the variation and reproducibility of the fiber bundles due to different sources of uncertainty and variation in the tractography input parameters.

Although these methods provide a first step to study uncertainty due to parameter settings, it remains a time consuming exploratory task for the user. This is especially true if parameters are correlated, and their interrelation needs to be investigated.



%% file: 03-2-uncertainty.tex
\subsection{Model Uncertainty and Selection}
\label{sec:model-uncertainty}

Methods for tractography that seek to recover more than a single fiber
direction in a given area have to make a judgement about how fiber
directions can be meaningfully recovered from the dMRI data.
The combination of measurement noise, partial voluming, and the
practical constraints on how many diffusion weighted images may be acquired create
uncertainty in the number of fibers present.
Qualitatively different than the angular uncertainty in a single fiber
direction, the traditional focus of probabilistic tractography, this
uncertainty can be a viewed as a kind of {\em model selection
  uncertainty}, which is described further in
Section~\ref{sec:evidence}.

Uncertainty in fiber number has been handled by different tests that
either statistically sample or deterministically choose a level of
model complexity (with an associated fiber number) from a nested set
of models.  Behrens et al.\ \cite{Behrens:2007} use Automatic
Relevance Determination (ARD) to probabilistically decide the number
of ``sticks'' (fibers) in their ball-and-multiple-stick model.
Within their probabilistic tractography, this achieves Bayesian Model
Averaging~\cite{GLK:Hoeting1999a} of the fiber orientation.
For deterministic tractography, Qazi et al.\ \cite{GLK:Qazi2009}
use a threshold on the (single, second-order) tensor planarity index
$c_p$~\cite{GLK:Westin1997} to determine whether to fit to the
diffusion weightes images a constrained two-tensor
model~\cite{GLK:Peled2006} that permits tracing two crossing fibers.
Schultz et al.\ compare different strategies for deciding the
appropriate number of fiber compartments, based on the diminishing
approximation error \cite{GLK:Schultz2008}, thresholding compartment
fraction coefficients of a multi-fiber model~\cite{Schultz:MICCAI10},
or by learning the number of fiber compartments using simulated data and
support vector regression~\cite{Schultz:MICCAI12}, which represents
uncertainty in the form of continuous estimates of fiber number.

Much of the work on determining the number of per-voxel fiber
components has been described outside of any particular tractography
method, but may nonetheless inform tractographic analysis.
Alexander et al.\ \cite{Alexander:2002} use an F-Test to find an 
appropriate order of Spherical Harmonic (SH) representation of the
ADC profile.
Jeurissen et al.\ \cite{Jeurissen:2010} decide the number of fibers
by counting significant maxima in the fiber orientation distribution
after applying the SH deconvolution (constrained by
positivity) of Tournier et al.\ \cite{Tournier:2007}.
The SH deconvolution of Tournier et al.\
\cite{Tournier:2007} in some sense involves model selection, because the
deconvolution kernel is modeled by the SH coefficients of the
voxels with the highest FA, presumably representing a single fiber.

Aside from the question of counting fibers, other work has examined
more broadly the question of which models of the diffusion weighted signal profile
are statistically supported.
Bretthorst et al.\ \cite{Bretthorst:2004} compute Bayesian evidence 
(see Section~\ref{sec:evidence}) to quantify the fitness of various
models of the diffusion weighted signal, producing maps of model complexity 
in a fixed baboon brain, and of evidence-weighted averages of 
per-model anisotropy.
Freidlin et al.\ \cite{Freidlin:2007} choose between the full
diffusion tensor and simpler constrained tensor models according to
the Bayesian information criterion (BIC) or sequential application of
the F-Test and either the t-Test or another F-Test.


%% file: 03-3-part-vol.tex
\subsection{Partial Voluming}
\label{sec:partial-voluming}


Tractography works best in voxels that contain homogeneously oriented
tissue. Unfortunately, many regions of the brain exhibit more complex
structures, where fibers cross, diverge, or differently oriented
fibers pass through the same voxel \cite{Alexander:2002}. This problem
is reduced at higher magnetic field strength, which affords increased
spatial resolution. However, even at the limit of what is technically
possible today
\cite{Heidemann:2011}, a gap of several orders of magnitude remains to
the scale of individual axons.


Super-resolution techniques combine multiple images to increase
effective resolution. Most such techniques use input images that are
slightly shifted with respect to each other and initial success has
been reported with transferring this idea to MRI
\cite{Peled:2001}. However, due to the fact that MR images are
typically acquired in Fourier space, spatial shifts do not correspond
to a change in the physical measurement, so it is unclear by which
mechanism repeated measurements should achieve more than an improved
signal-to-noise ratio \cite{Scheffler:2002,Peled:2002}. It is less
controversial to compute images that are super-resolved in
slice-select direction \cite{Greenspan:2002,Scherrer:2011} or to
estimate fiber model parameters at increased resolution via smoothness
constraints \cite{Nedjati:2008}.

Track density imaging \cite{Calamante:2010} uses tractography to
create super-resolved images from diffusion MRI. After randomly
seeding a large number of fibers, the local streamline density is
visualized. It is computed by counting the number of lines that run
through each element of a voxel grid whose resolution can be much
higher than during MR acquisition. Visually, the results resemble
those of line integral convolution, which had been applied to dMRI
early on \cite{Hsu:2001,Zheng:2003}.


%% file: 04-1-evidence.tex
\subsection{Evidence for Model Selection}
\label{sec:evidence}

Many of the methods for finding per-voxel fiber count
(or more generally the per-voxel signal model)
described in Section~\ref{sec:model-uncertainty}
share two notable properties which may
be reconsidered and relaxed in future research.
First, they deterministically calculate the single best
model, with hard transitions between the regions best explained by one model versus
another~\cite{Alexander:2002,Freidlin:2007,GLK:Qazi2009,GLK:Schultz2008,Schultz:MICCAI10,Jeurissen:2010}.
Yet we know that partial voluming (Sect.~\ref{sec:partial-voluming})
creates smooth transitions between different neuroanatomic tissue regions.
Though computational expensive, Markov Chain Monte Carlo (MCMC)
sampling of both model parameter space and the set of models enables
averaging information from more than one
model~\cite{Behrens:2007,Bretthorst:2004}.
Second, most methods work within a particular hierarchical set
of linearly ordered models (SH of different orders~\cite{Alexander:2002},
ball and multiple sticks~\cite{Behrens:2007}, sum of higher-order
rank-1 terms~\cite{Schultz:MICCAI10}).
One can easily imagine configurations, however, that confound such a 
linear ordering: an equal mix of two fibers crossing and isotropic
diffusion (perhaps due to edema), or a mix of one strong fiber
and two weaker equal-strength fibers.
Furthermore, there is rarely objective comparison or reconciliation
between disjoint sets of models.

An informative perspective on these situations may be gained by
directly visualizing, either on data slices or by some form of
volume rendering, the fitness of a large palette of possible models.
In a Bayesian setting, the model fitness can be quantified by the
marginal likelihood of the data $\mathbf{x}$ given the model $M_k$, or
the model {\em evidence}, computed by integrating over the model
parameter space $\bm{\theta}_k$ \cite{GLK:Kass1995}.
\begin{align}
\underbrace{P(\mathbf{x}|M_k)}_{\mbox{evidence}} &= \int
\underbrace{P(\mathbf{x}|\bm{\theta}_k,M_k)}_{\mbox{likelihood}}
\underbrace{P(\bm{\theta}_k|M_k)}_{\mbox{prior}}
d\bm{\theta}.
\end{align}
Bretthorst et al.\ \cite{Bretthorst:2004} have pioneered the
calculation and visualization of model evidence for dMRI, but many 
possible directions are left unexplored, including the application
to counting fibers, and to models that account for intra-voxel
fanning or bending~\cite{GLK:NedjatiGilani2009}.


%% file: 04-2-reproducibility.tex
\subsection{Reproducibility, Seeding, and Preprocessing}
\label{sec:preprocessing}

The reproducibility of tractography depends on many factors. The
manual placement of seed points is an obvious concern. Detailed
written instructions improve reproducibility between operators
\cite{Ciccarelli:2003}, especially across sites
\cite{Wakana:2007}. Combining multiple seed regions with logical
operators makes the results more reproducible
\cite{Huang:2004,Heiervang:2006} and seeding protocols for up to
11~major fiber bundles have been developed this way
\cite{Wakana:2007}. Warping individual brains to a standard template
has also been reported to increase reproducibility
\cite{Heiervang:2006,Vollmar:2010}. Selecting streamlines from a whole
brain tractography via three-dimensional regions of interest
\cite{Blaas:2005} or semi-automated clustering \cite{Voineskos:2009}
is an alternative way to reproducibly extract fiber bundles.

When the same person places the seeds on a repeated scan, the
resulting variability is generally higher than when different
observers follow a written protocol to place seeds in the same data
\cite{Ciccarelli:2003}. Within the same session, measurement noise is
the main limiting factor \cite{Ding:2003}. Between sessions,
differences in exact head positioning and other subtle factors
increase the variability noticably \cite{Vollmar:2010}.

Reproducibility suffers even more when repeating the measurement on a
different scanner \cite{Pfefferbaum:2003}. Even a pair of nominally
identical machines has produced a statistically significant bias in
Fractional Anisotropy \cite{Vollmar:2010}. Improving calibration
between sessions or scanners via software-based post-processing
appears possible \cite{Vollmar:2010}, but has not been widely explored
so far. 

More time consuming measurement protocols generally afford better
reproducibility. Even though Heiervang et al.\ \cite{Heiervang:2006}
report that the improvement when using 60 rather than 12 gradient
directions was not statistically significant, Tensouti et al.\
\cite{Tensaouti:2011} report a clear improvement between 6 and
15~directions, which continues -- at a reduced rate -- when going to
32~directions. Farrel et al.\ \cite{Farrell:2007} use 30~directions
and demonstrate a clear improvement when averaging repeated measurements.

Finally, reproducibility depends on the tractography algorithm
\cite{Tensaouti:2011}, its exact implementation \cite{Burgel:2009}, as
well as the methods used for pre-processing the data
\cite{Jones:2010NMR,Vollmar:2010} and their parameter settings. Given
that the reproducibility of tractography will be crucial for its wider
acceptance in science and medicine, more work is needed that
specifically targets these problems.


%% file: 05-conclusion.tex


Reproducibility of dMRI tractography is a fundamental problem that limits the acceptance of this technique in clinical practice and neuroscience research. Although some effort has been made to include uncertainty information in the tractography results, several open issues remain that need further investigation.

Probabilistic tractography is established, but visualization research has concentrated on deterministic streamline-based techniques, and few techniques have been developed to visualize the information obtained by probabilistic methods.
There are several sources of uncertainty in the tractography visualization pipeline. However, only a few of them have been explored, and if at all studied, they are often considered independently with no connection to each other. Techniques that investigate the impact of parameters on the fiber tracking results and that aim to reduce the impact of user bias through parameter selection have been investigated only recently. Model selection and data preprocessing have hardly been studied with respect to their effects on tractography results.

Techniques that allow the combined analysis of uncertainty from different sources in the same framework, and that facilitate an understanding of their influence on the final tractography result are still missing. One challenge faced by visualization systems that aim to aid understanding of these uncertainties is to display this additional information efficiently and effectively, without causing visual clutter.

Ultimately, uncertainty visualization should contribute to making fiber tracking a more reliable tool for neuroscience research, and to conveying the information needed for the decision making process in clinical practice.